 \documentclass[final,5p,times,twocolumn,authoryear]{elsarticle}

\usepackage{amssymb}
\usepackage{lipsum}
\usepackage{multirow} 
\usepackage{booktabs}
\usepackage{stfloats}
\usepackage {amsmath}
\usepackage{subcaption}

\journal{Biomedical Signal Processing and Control}

\begin{document}

\begin{frontmatter}



\title{Optimizing Drug Delivery in Smart Pharmacies: A Novel Framework of Multi-Stage Grasping Network Combined with Adaptive Robotics Mechanism}


\author[First,Second]{Rui Tang}\author[Third]{Shirong Guo}\author[Third]{Yuhang Qiu}\author[Fourth]{Honghui Chen}\author[Second]{Lujin Huang}\author[Third]{Ming Yong} \author[Fifth]{Linfu Zhou}\author[First]{Liquan Guo*}

\affiliation[First]{organization={Suzhou Institute of Biomedical Engineering and Technology, Chinese Academy of Science},
            postcode={215163}, 
            state={Suzhou},
            country={China}}
\affiliation[Second]{organization={Department of Electronic Information Engineering, Fuzhou University},
            postcode={350108}, 
            state={Fuzhou},
            country={China}}
\affiliation[Third]{organization={Faculty of Engineering, Monash University},
            postcode={Victoria 3800}, 
            state={Clayton},
            country={Australia}}
\affiliation[Fourth]{organization={Department of Physics and Information Engineering, Fuzhou University},
            postcode={350108}, 
            state={Fuzhou},
            country={China}}
            
\cortext[cor1]{Corresponding author} 
\ead{guolq@sibet.ac.cn}

\affiliation[Fifth]{organization={Department of Respiratory and Critical Care Medicine, The First Affiliated Hospital, Nanjing Medical University},
            postcode={211166}, 
            state={Nanjing},
            country={China}}

\begin{abstract}
Robots-based smart pharmacies are essential for modern healthcare systems, enabling efficient drug delivery. However, a critical challenge exists in the robotic handling of drugs with varying shapes and overlapping positions, which previous studies have not adequately addressed. To enhance the robotic arm's ability to grasp chaotic, overlapping, and variously shaped drugs, this paper proposed a novel framework combining a multi-stage grasping network with an adaptive robotics mechanism. The framework first preprocessed images using an improved Super-Resolution Convolutional Neural Network (SRCNN) algorithm, and then employed the proposed YOLOv5+E-A-SPPFCSPC+BIFPNC (YOLO-EASB) instance segmentation algorithm for precise drug segmentation. The most suitable drugs for grasping can be determined by assessing the completeness of the segmentation masks. Then, these segmented drugs were processed by our improved Adaptive Feature Fusion and Grasp-Aware Network (IAFFGA-Net) with the optimized loss function, which ensures accurate picking actions even in complex environments. To control the robot grasping, a time-optimal robotic arm trajectory planning algorithm that combines an improved ant colony algorithm with 3-5-3 interpolation was developed, further improving efficiency while ensuring smooth trajectories. Finally, this system was implemented and validated within an adaptive collaborative robot setup, which dynamically adjusts to different production environments and task requirements.  Experimental results demonstrate the superiority of our multi-stage grasping network in optimizing smart pharmacy operations, while also showcasing its remarkable adaptability and effectiveness in practical applications. 
\end{abstract}




\end{frontmatter}




\section{Introduction}
\label{introduction}
In the modern healthcare system, the dramatic increase in the number of patients and the rapid expansion of healthcare needs have made smart pharmacies a critical solution to the challenges faced by traditional pharmacies \citep{raza2022artificial, m2019critical}. Despite the numerous benefits brought by the introduction of smart pharmacies \citep{rajpurkar2022ai}, several issues still need to be addressed \citep{2023SPIE12702E..04W}. One key challenge is improving accuracy and efficiency in the drug distribution. Despite the adoption of advanced technologies and automation equipment in smart pharmacies, including the use of robotic arms for medication handling, these robotic systems are primarily limited to grasping medications of singular shapes and types. They are unable to adapt to the complex and variable conditions of the dispensing process. Consequently, pharmacies continue to rely heavily on manual dispensing. This reliance leads to heavy manual tasks, misdistribution of medicines, and inefficient pick-up procedures\citep{khatib2020robotic}. These issues not only affect the patient’s medical experience but also increase the workload of healthcare workers and reduce the overall efficiency of healthcare services. Therefore, developing an accurate sorting system suitable for complex drug scenarios is crucial for advancing automated drug management in smart healthcare \citep{soori2023artificial}.

Most smart pharmacies currently use the slanting slot positioning method for dispensing medicine \citep{lin2020evaluation}. This method often results in drugs being randomly positioned, overlapping each other, and presenting varied grasp angles. Additionally, the variation in the number and shape of drugs dispensed each time poses significant challenges for robotic grasp detection. Early smart pharmacy robots, constrained by simpler mechanized operations and barcode recognition technology \citep{csencan2019general}, could only perform pre-programmed grasping tasks on fixed medication shelves, lacking the ability to adapt to complex environments. The introduction of computer vision technology, however, has revolutionized this field. High-resolution cameras, combined with advanced image processing and machine learning algorithms, now enable robots to recognize the packaging features of medicines directly through vision, greatly enhancing the flexibility and accuracy of recognition \citep{yan2021design, pratheep2022design}. Despite these advances, performance remains limited in particularly complex or cluttered scenarios, such as the random arrangement and overlapping of medicines in a smart pharmacy.

Therefore, in order to meet the challenges of grasping drugs with chaotic overlap and different shapes and postures in the actual dispensing environment of intelligent pharmacies. This paper proposes a system for intelligent drug distribution that integrates a multi-stage grasping framework. 
The proposed framework initiates by preprocessing images with an enhanced Super-Resolution Convolutional Neural Network (SRCNN) algorithm. Subsequently, the YOLOv5+E-A-SPPFCSPC+BIFPNC (YOLO-EASB) instance segmentation algorithm is employed to accurately segment each drug in the preprocessed images. By evaluating the completeness of the segmentation masks, the framework identifies the most suitable drugs for grasping. Then, these segmented drugs were processed using Improved Adaptive Feature Fusion and Grasp-Aware Network (IAFFGA-Net), enabling precise grasping even in complex environments. To control the robotic arm, a trajectory planning framework for time-optimal trajectory optimization is developed using an improved Particle Swarm Optimization (PSO) algorithm. This framework constructs a continuous trajectory using 3-5-3 segmented polynomials, interpolating the trajectory while adhering to the robotic arm's speed and acceleration constraints. To enhance search capabilities and mitigate early convergence and late optimization issues typical in standard particle swarm algorithms, a sequence generation strategy based on Logistic-Tent dual chaos mapping is introduced for particle swarm initialization. Chaotic perturbation techniques are applied to enable early-ripening particles to escape local extremes. Finally, an adaptive collaborative robot system is implemented and validated, demonstrating its ability to dynamically adjust to various production environments and task requirements. Experimental results underscore the superiority of this multi-stage grasping network in optimizing smart pharmacy operations.

The main contributions of our work are as follows:

(1) The novel multi-stage grasping framework preprocesses images using an improved Super-Resolution Convolutional Neural Network (SRCNN), enhancing image quality and model prediction speed. The preprocessed images are fed into the YOLO-EASB instance segmentation algorithm, which refines the YOLOv5 spp and FPN structures and introduces an adaptive dual attention mechanism for accurate drug segmentation in chaotic occlusions. The integrity of the segmentation mask is evaluated to identify the optimal drug for grasping, and the IAFFGA-Net ensures precise grasping.

(2)  The time-optimal robotic arm trajectory planning algorithm combines the ant colony algorithm with 3-5-3 interpolation planning. It introduces dual chaos Logit-Tent mapping for particle initialization and employs a nonlinear decreasing strategy to adjust inertia weight, addressing premature and slow convergence issues in standard particle swarm algorithms.

(3) The implementation and verification of the drug distribution robot system can accurately grasp drugs of different shapes and types in chaotic and obstructed environments, which is closest to the real intelligent pharmacy drug distribution environment. It has shown significant adaptability and effectiveness in practical applications.

\section{Related work}
\subsection{Pharmacy automation system}
With the increasing population and the growing burden on hospitals, the demand for robot-assisted pharmacies has become more urgent. The Iron-1200 automated dispensing machine, which does not employ robotic arms, adopts a semi-automatic loading method and is equipped with a manual operation panel and mechanical devices, achieving an hourly stocking rate of 1,500 boxes. However, this method requires operators to organize and input information in advance \citep{jin2017ant}. For methods utilizing robotic arms for dispensing,Liu et al. combined image and text information to accurately grasp specific objects (bottles), but this approach is unable to handle other types of pharmaceuticals with different shapes\citep{liu2022scene}. Ren et al. used the Fast-RCNN algorithm for detection and point cloud matching to grasp medication boxes, but this method only supports the grasping of single objects in a scene and cannot be applied to the chaotic environments of real-world pharmacy sorting  \citep{ren2016faster, zou2022efficient}. These intelligent pharmacy systems achieve a certain degree of automation but are limited by algorithms and specific scenarios, making them unsuitable for the complex distribution environments of real pharmacies.
\subsection{Grasp detection}
Grasp detection in complex occlusion environments is a critical and challenging research area in robotics. Existing methods primarily focus on the prediction of the grasping tool's pose and the estimation of the object's pose. Grasp frame prediction methods typically concentrate on the posture of the grasping tool rather than the object's pose. For instance, the studies by Lenz et al. predict grasping frames directly from images without estimating the objects' poses. While effective for scenes with simple and dispersed objects, these methods are less capable of predicting the optimal grasp pose in stacked environments \citep{ren2016faster,lenz2015deep}. Posture estimation methods employing deep learning techniques aim to predict an object's six-degree-of-freedom pose using neural networks. Examples include the works of \citep{bukschat2020efficientpose,9157169}. However, these approaches do not fundamentally address the occlusion problem inherent in grasping tasks. Several studies have proposed methods to handle cluttered scenes. Cheng et al. use a boundary band method and topological ordering to establish the depth order of overlapping instances. However, this approach requires specific constraints, is computationally intensive, and has slow inference speeds \citep{cheng2010repfinder}. Zhan et al. employ a self-supervised segmentation network to recover the complete structure of an object and determine the occlusion relationships between neighboring objects, yet this method fails to detect the object's class \citep{zhan2020self}. Ainetter et al. leverage semantic segmentation results to provide auxiliary cues to the grasping network, thereby improving grasp detection accuracy. However, this method performs well only under conditions of minimal occlusion and clutter, struggling with heavy occlusions and complex environments \citep{ainetter2021end}. In summary, existing methods often struggle with complex, stacked, and occluded scenes. Thus, developing a grasping framework with adaptive processing capabilities to manage complex occlusion environments remains essential.

\subsection{Trajectory planning}
In the field of robot motion control, early trajectory planning primarily relied on fundamental methods, utilizing mathematical tools such as lines, arcs, and advanced curves for geometric construction. Boryga et al. employed high-order polynomials (5th, 7th, and 9th) for joint spatial trajectory planning of robotic arms. While these methods ensure continuous velocity, acceleration, and jerk, they suffer from the drawbacks of high-order polynomial characteristics and poor convex envelope properties \citep{boryga2009planning}. To overcome the limitations of single-order polynomials in trajectory planning, Dincer et al.combined third-order polynomials and Bessel curves to achieve smooth trajectories at the start and end points, with better convergence at the path points \citep{dinccer2019improved}. Ming et al. utilized a 3-5-3 piecewise polynomial function to interpolate the motion trajectory of robotic arms, achieving high-precision and smooth motion \citep{ming2018improved} . With technological advancements and the increasing complexity of application scenarios, simple trajectory fitting no longer meets the multidimensional optimization requirements for efficiency and energy utilization. In this context, optimal trajectory planning that integrates optimization algorithms with basic trajectory construction demonstrates superior performance. This approach aims to find the optimal trajectory by modeling and solving specific objective functions under given constraint conditions. Qiao et al. employed an improved genetic algorithm for trajectory planning using fifth-degree polynomial interpolation, achieving global optimization of joint trajectories. However, the use of fifth-degree polynomials in their interpolation resulted in longer computational times compared to piecewise polynomial interpolation \citep{qiao2020trajectory}. Liu et al. utilized an improved particle swarm optimization algorithm combined with 4-3-4 polynomial trajectories for time optimization, introducing dynamic learning factors. Despite this, the method's multiple segmentation times and high computational complexity remain significant drawbacks \citep{liu2020improved} . In summary, while significant progress has been made in trajectory planning research, challenges such as high computational complexity and time consumption persist. Addressing these issues is crucial for advancing the efficiency and effectiveness of robot motion control systems.

\section{Method}
\subsection{Framework of Multi-Stage Grasping Network Combined with Adaptive Robotics Mechanism }
\begin{figure*}[h]
    \centering               
\includegraphics[width=1.0\textwidth]{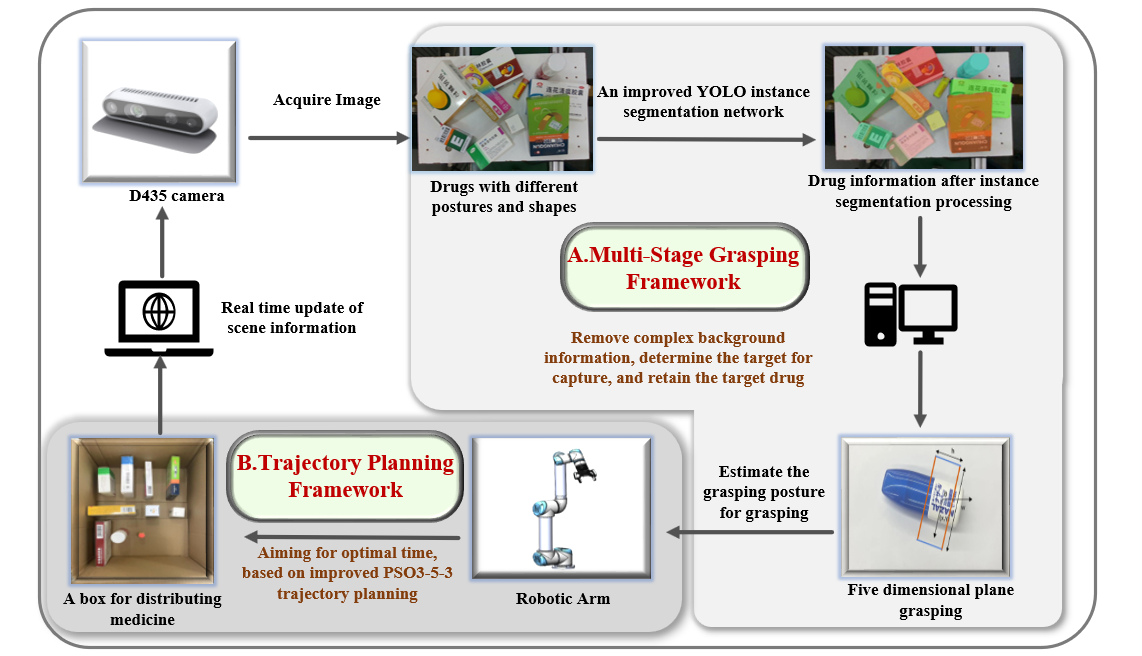}
    \caption{The framework of the Multi-Stage Grasping Network Combined with Adaptive Robotics Mechanism} 
    \label{fig.1}
\end{figure*}

As shown in Figure \ref{fig.1}, this paper proposes a comprehensive grasping system architecture that integrates the SRCNN improvement algorithm, the YOLO-EASB, and the IAFFGA-Net to achieve precise grasping of medicine boxes. Firstly, a D435 binocular camera captures the chaotic and overlapping medicines. Then, a multi-stage stage grasping network framework grasps the medicines. The robotic arm trajectory is dynamically planned using an optimized particle swarm algorithm, enabling the grasping network to find the optimal path to the grasping position within a limited time and minimize potential risks during the process. In summary, through the synergistic effect of multi-stage technology, the proposed grasping detection system architecture successfully achieves accurate, efficient, and stable grasping operations of drugs in complex environments.
\subsection{Multi-Stage Grasping Network Framework}

\begin{figure*}[t!]
    \centering               
\includegraphics[width=.90\textwidth]{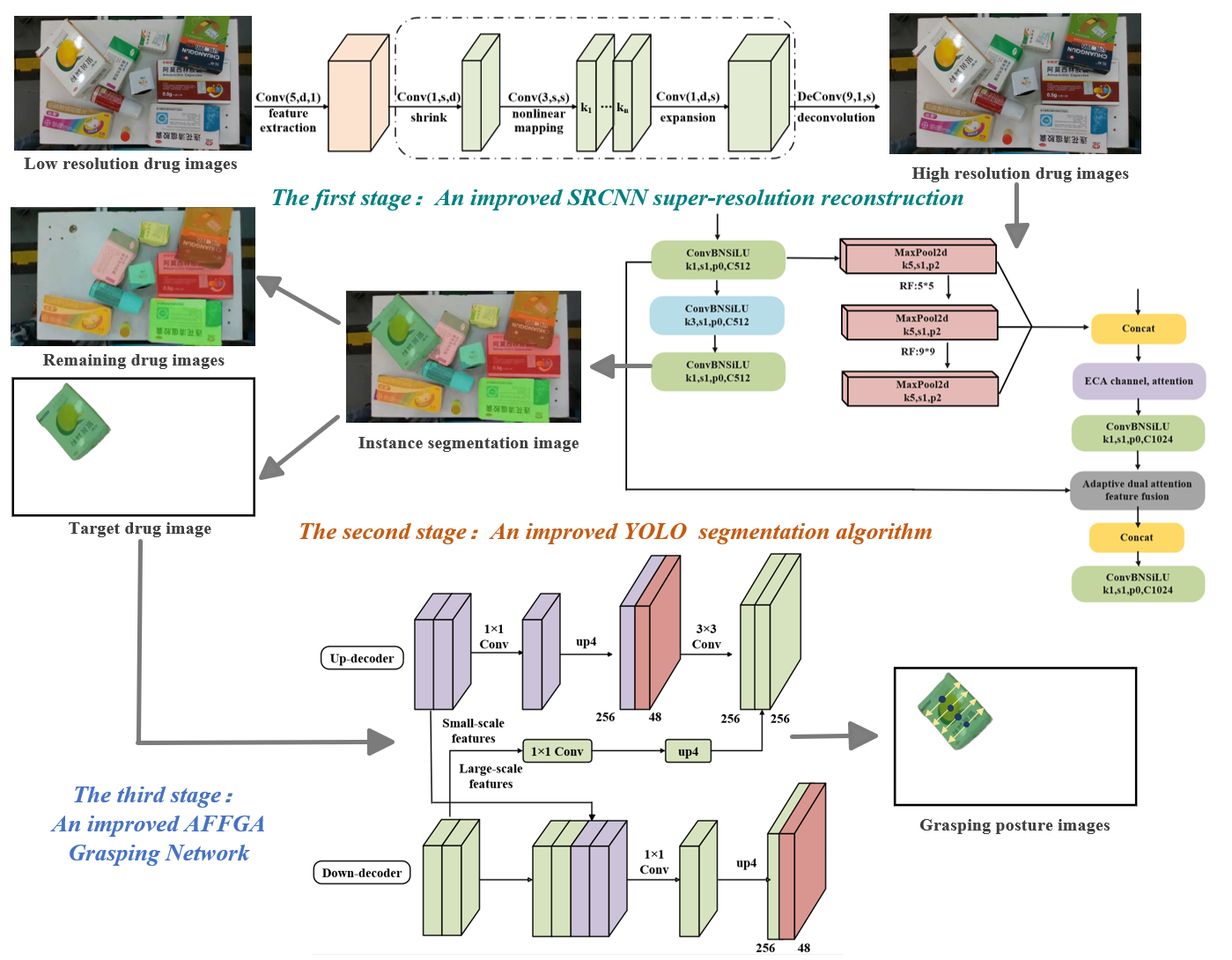}
    \caption{Multi-Stage Grasping Framework Network Architecture}
    \label{fig.2}
\end{figure*}

The AFFGA-Net \citep{9586564} consumes significant computational resources during detection, and complex background information can reduce its accuracy. To address this, the original image is first reconstructed using hyper-segmentation, and then the high-resolution image is input into the YOLO-EASB. This process eliminates complex background information and extracts the outline of objects to be grasped. The mask of the target object is then input into the IAFFGA-Net as an intermediate image. This approach significantly enhances computational speed and reduces the environmental influence on the target, thereby improving grasping accuracy. The network architecture is shown in Figure \ref{fig.2}.
\subsubsection{Image super-segmentation preprocessing based on improved SRCNN}
\begin{figure}[h]              
\includegraphics[width=0.5\textwidth]{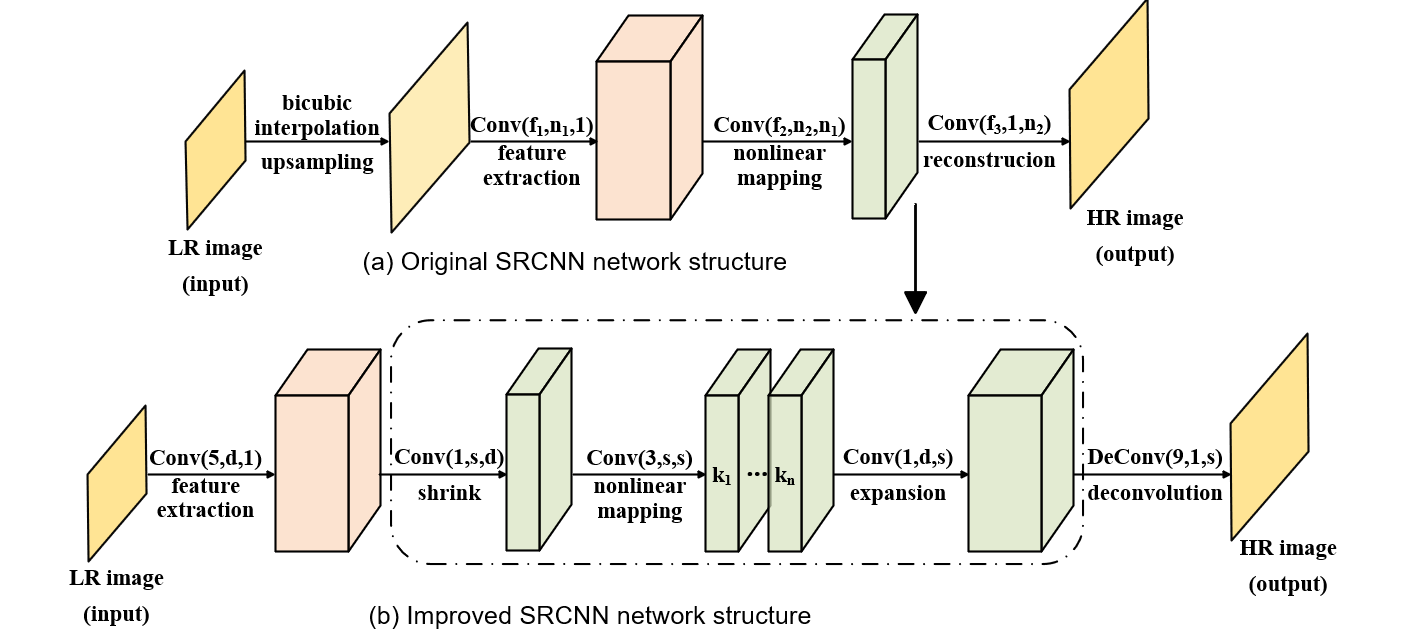}
    \caption{Flowchart of Hyperpartition Reconstruction Algorithm(a)Original SRCNN network structure(b)Improved SRCNN network structure}
    \label{fig.3}
\end{figure}

\textbf{a. SRCNN algorithm}\\
Super-resolution reconstruction of drug images can effectively improve the accuracy of subsequent drug segmentation. The deep learning-based SRCNN super-resolution reconstruction method \citep{liu2020deep} consists of four main parts: up-sampling, feature extraction and representation, nonlinear mapping, and reconstruction. As shown in Figure \ref{fig.3}(a), the method first preprocesses the original low-resolution (LR) image using the bicubic interpolation algorithm, enlarging the scale of the original input image to a set scale. It then performs a convolution operation on the enlarged LR image to obtain a high-dimensional ($n_1$-dimensional) feature vector representation. This $n_1$-dimensional feature vector is mapped to an $n_2$-dimensional feature vector for reconstructed high-resolution feature representation. Finally, the reconstructed high-resolution (HR) image is obtained by averaging the predicted overlapping high-resolution blocks. Compared with traditional modeling methods, SRCNN achieves higher reconstructed image quality but depends on image region information and has a slower inference speed.\\

\textbf{b. Improved SRCNN algorithm}\\
To meet the demand for real-time performance while acquiring high-resolution images, an improved SRCNN algorithm is proposed. Instead of performing bicubic interpolation on the input image, the original low-resolution image is used as input, and an inverse convolutional layer is introduced at the end of the network for up-sampling. Additionally, the nonlinear mapping layer in SRCNN is replaced with contraction, mapping, and expansion operations, utilizing smaller convolutional kernels and a deeper network structure. This approach converges more effectively than the original model, resulting in lower error while maintaining high reconstruction quality. The flow of the improved algorithm is shown in Figure \ref{fig.3}(b).

This paper defines the convolutional layer as DeConv($f_i$, $n_i$, $c_i$) and the inverse convolutional layer as $f_i$, $n_i$, $c_i$, where $f_i$, $n_i$, and $c_i$ represent the convolutional kernel size, the number of convolutional kernels, and the number of channels, respectively. 
The feature extraction layer uses a 5×5 convolutional kernel for the original low-resolution image instead of the 9×9 kernel in SRCNN for feature extraction, with the number of channels ($c_i$) as 1 and the number of convolution kernels ($n_i$) as d. The shrinkage layer employs 1×1 convolutional kernels to reduce the LR feature dimension from d to s with a smaller number of convolutional kernels (s). The nonlinear mapping layer replaces the original wide convolutional kernels with s kernels of size 3×3, maintaining the number of channels ($c_i$) as s. The expansion layer is added after the nonlinear mapping to expand the HR feature dimension, using s convolutional kernels of size 1×1 for dilation to maintain consistency with the shrinkage layer. The inverse convolution layer performs the inverse process of convolution, and with a step size of k, the resolution is improved by k times, using a 9×9 inverse convolution kernel. By refining these layers, the model achieves higher computational efficiency and better image reconstruction quality.

After the above improvements, our SRCNN module not only enhances the clarity of low-resolution images and ensures the accuracy of drug segmentation but also significantly reduces computational load and memory consumption. This improves overall efficiency while maintaining detection accuracy.
\subsubsection{YOLO-EASB instance segmentation model}
The varying shapes, significant scale differences, and complex background environments of drugs lead to low accuracy in drug instance segmentation. To address this, an improved YOLOv5-based drug instance segmentation method is proposed. The following sections detail the enhancements to the SPP module and the construction of the BiFPN cross-scale feature fusion network in YOLOv5.\\
\textbf{a. E-A-SPPFCSPC}\\
Processing drugs of varying scales can make it difficult for models to capture detailed information and can slow detection speed due to increased computation. To enhance the model's ability to handle targets at different scales while reducing computational effort and improving detection speed, the SPPF module in YOLOv5 is replaced with the SPPFCSPC module\citep{wang2022yolov7}, which includes Spatial Pyramid Pooling, Shortcut, and Cross-Stage Partial Network (CSPNet).Based on the multiscale features generated by the SPPFCSPC module, the Efficient Channel Attention (ECA) and the proposed Adaptive Dual Attention Feature Fusion (ADaFF) modules are incorporated. These modules enable the model to adaptively assign response weights to each channel, highlighting key features in non-obscured regions during partial occlusion. They efficiently integrate features across layers and suppress redundancy, allowing the network to distill high-level abstract features from the deep network while fully utilizing delicate texture information from shallow inputs.
\begin{figure}[h]
    \centering               
\includegraphics[width=0.5\textwidth]{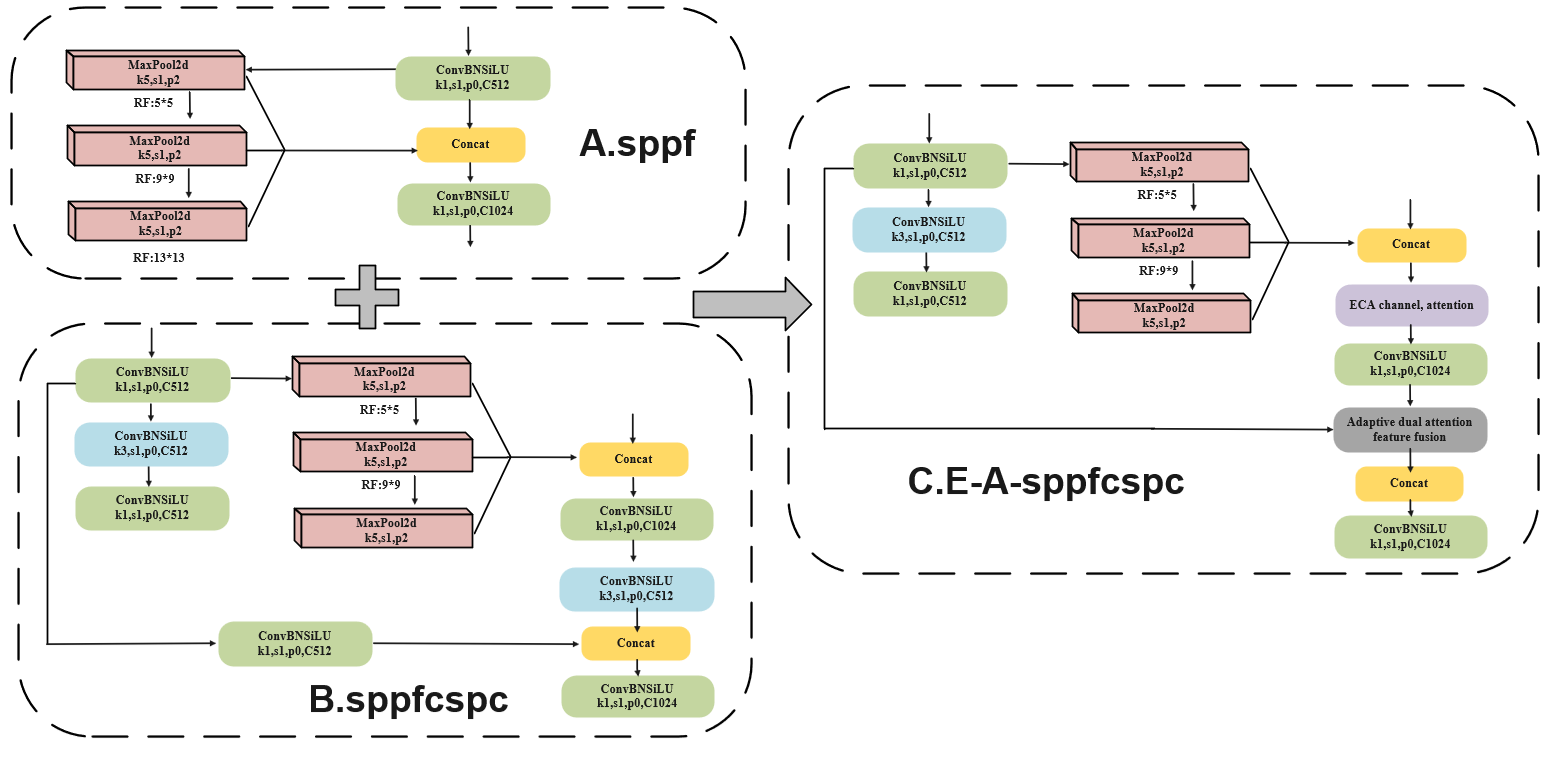}
    \caption{The architecture of E-A-SPPFCSPC Module}
    \label{fig.4}
\end{figure}

The ECA Attention Module is based on the attention mechanism of Squeeze and Excitation Networks (SENet), removing the fully-connected layer and introducing a new computational structure.This structure utilizes adaptive one-dimensional convolution to learn dependencies between neighboring channels, enhancing information interaction without significantly increasing parameters. The specific process is provided as follows: 

For the input feature map $\mathrm{X} \in \mathrm{R}^{\mathrm{H} \times \mathrm{W} \times \mathrm{C}}$, a global average pooling operation is applied to transform the features of each channel into a global descriptor of size 1×1×C. The size of the one-dimensional convolution kernel, \( k \), is determined using equation (1), based on the number of input channels \( C \). In this paper, the parameter \( \gamma \) = 1.5, and \( b \) = 1. $\varphi(C)$ selects \( k \) as the closest singular value to the absolute value computed, representing the number of neighboring channels each channel needs to capture dependencies from. 
Finally, the weight \( \omega_i \) for each channel is obtained using the Sigmoid function,and $\sigma$ denotes the Sigmoid function. This weight is then multiplied with each channel of the input feature map \( X \) to re-weight the channel features accordingly.

\begin{figure}[h]
    \centering               
\includegraphics[width=0.5\textwidth]{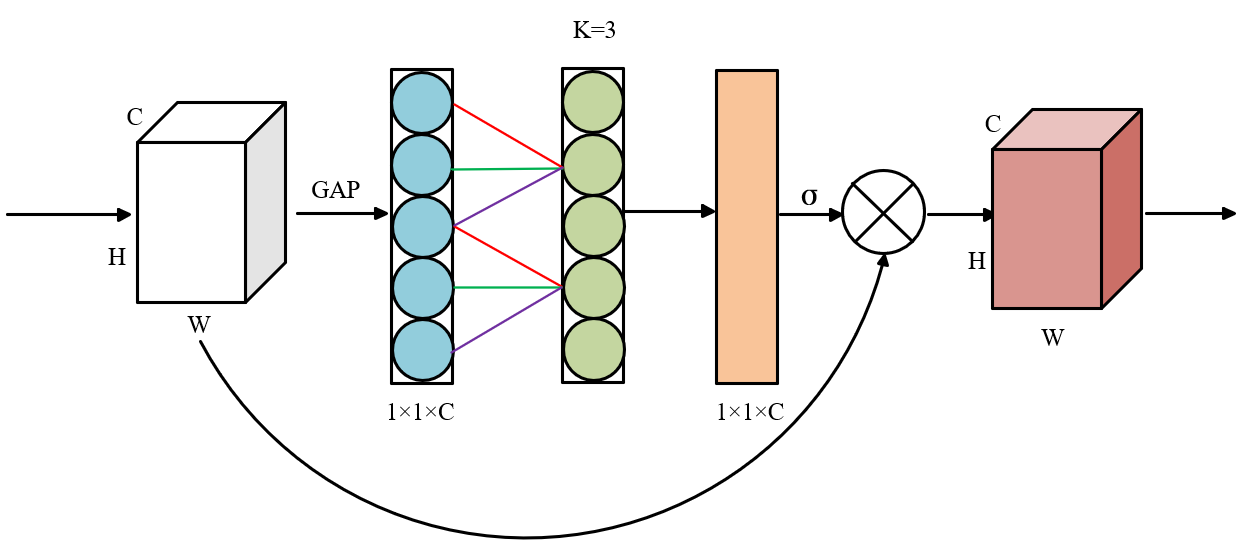}
    \caption{The architecture of ECA Module}
    \label{fig.5}
\end{figure}

\begin{figure}[h]
    \centering               
\includegraphics[width=0.5\textwidth]{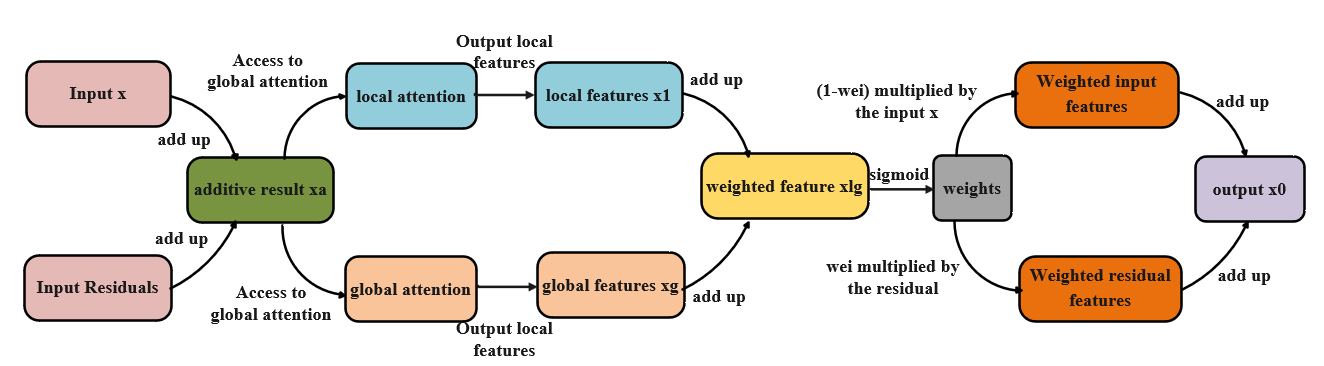}
    \caption{The architecture of ADaFF Module}
    \label{fig.6}
\end{figure}

The weights are calculated as shown in equation (1)(2):
\begin{equation}
\mathrm{k}=\varphi(C)=\left|\frac{\log _2(C)}{\gamma}+\frac{b}{\gamma}\right|_{\text {odd }}
\end{equation}
\begin{equation}
\omega_{\mathrm{i}}=\sigma\left(\sum_{j=1}^{\mathrm{k}} \omega_i^j y_i^j\right), y_i^j \in \Omega_i^k
\end{equation}
where $\Omega_i^k$ represents the set of k neighboring channels. 
The principle of ECA structure is shown in Figure \ref{fig.5}.\\
The ADaFF module is proposed in this study to improve the representation capability of features. This module sums the input features with the residual features channel-by-channel and then processes them through two sub-processes: local attention and global attention. The module structure is depicted in Figure \ref{fig.6}. The local attention mechanism utilizes the Efficient Channel Attention (ECA) mechanism to enhance the representation of local features, while global attention captures global contextual information to enhance feature integrity. The output features from these attention mechanisms interact through a feature fusion strategy to achieve adaptive integration. Finally, in the output stage, local and global features are weighted and independently calculated to generate the final output features.

\textbf{b. BiFPNC}\\
The original YOLOv5 feature fusion network uses an FPN \citep{lin2017feature} + PAN structure, which cascades feature maps to the same resolution, limiting the full utilization of feature information. To address the issue of drug targets with diverse classes, large size variations, and complex shapes, this paper improves upon the FPN + PAN feature fusion network by introducing BiFPN \citep{wang2024measurement}. This enhancement aims to achieve a more comprehensive interaction between deep semantic information and shallow spatial information.

\begin{figure}[h]
    \centering               
\includegraphics[width=0.5\textwidth]{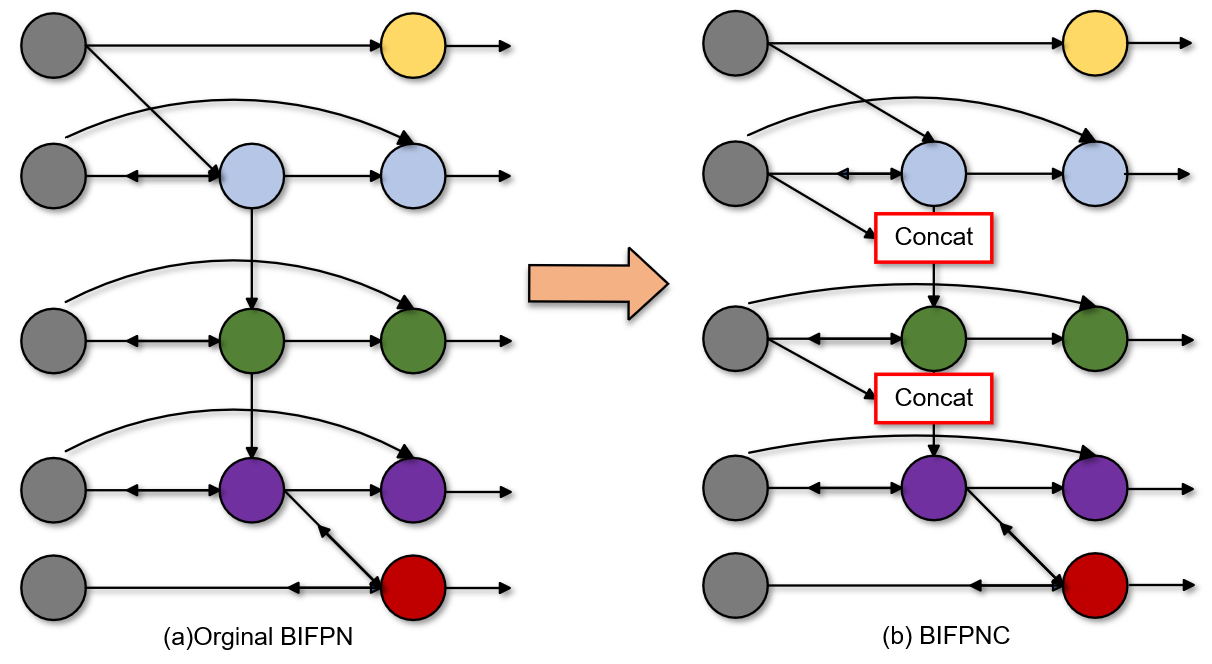}
    \caption{The architecture of BIFPNC Module}
    \label{fig.7}
\end{figure}

As shown in Figure \ref{fig.7}(a), the original BiFPN introduces learning parameters to express the weights of each input feature map. This mechanism enables the model to learn which feature layers are more important during training, allowing for efficient feature fusion by normalizing these weights. However, BiFPN tends to prioritize the selection of upper layers in feature weight selection, which may cause the fusion process to overlook important information from shallow features, especially fine-grained details crucial for accurate object detection and segmentation.

Therefore, this paper proposes a strategy to enhance feature fusion through jump connections, as shown in Figure \ref{fig.7}(b). By introducing shallow features directly into deep feature fusion, the network not only takes into account features from the previous layer but also from the initial layer, thereby utilizing high-level semantic information and low-level spatial information. By concatenating shallow features with deep features normalized by weights, shallow features are effectively retained, extending the representational ability of the feature map. This allows the network to capture the details of the target drug over a larger range, improving the perception of boundaries and textures. \\

The principle is shown in equations (3) to (7), where weight normalization is performed first, followed by weighted feature fusion, activation after weighted feature fusion, and finally feature splicing and convolution operations.\\
\begin{equation}
w^{\prime}=\frac{w}{\sum w+\epsilon}
\end{equation}
\begin{equation}
y=w_0^{\prime} \cdot x_0+w_1^{\prime} \cdot x_1+w_2^{\prime} \cdot x_2
\end{equation}
\begin{equation}
y^{\prime}=\operatorname{SiLU}(y)
\end{equation}
\begin{equation}
z=\operatorname{concat}\left(x_0, y^{\prime}\right)
\end{equation}
\begin{equation}
\text { output }=\text { Conv } 2 d(z)
\end{equation}
where $w^{\prime}$ represents the weights to be solved, weights [w0, w1, w2] are the learnable parameters obtained by model training, and x = [x0, x1, x2] are the feature maps of different scales. $\epsilon$ is 0.0001 to ensure stability. Output denotes the output features, Conv denotes the convolutional transform, $SiLU$ is the activation function, $y$ is the fused feature map, $y^{\prime}$ is the activated feature map, concat is the feature splicing, and z is the spliced feature map.

\subsubsection{Grasping selection}
The instance segmentation network can detect occlusion and the stacking of medicines within the scene but cannot directly infer their stacking relationship. To determine this relationship, each drug's mask is evaluated based on its completeness. A fully visible drug is assigned a complete mask, while partially occluded drugs are given partial masks. The completeness of the segmentation mask is then used to calculate a grasp score X, X for each object, factoring in parameters such as visibility, edge clarity, and overlap with neighboring objects. The object with the highest grasp score is identified as the first target for grasping. This object is then segmented separately from the scene, while the remaining drugs are considered background and processed accordingly before being fed into the subsequent IAFFGA-Net for further grasping actions. This approach ensures that the most accessible object is prioritized for accurate and efficient grasping, even in complex stacking scenarios.

\subsubsection{AFFGA-Netmwith the optimized loss function}
The AFFGA-Net\citep{wang2021high} designs Oriented Arrow Representation (OAR) models to represent parallel gripping jaws and simplified three-finger gripping jaw configurations, enhancing adaptability to objects of different sizes and shapes. The OAR model is predicted at each pixel point on the image to precisely describe potential grasping poses. The Adaptive Grasping Attribute Model (AGA-model) \citep{wang2021high} adaptively represents an object's grasping attributes, eliminating conflicting grasping angles and simplifying training by merging OAR models on neighboring pixel points. In the adaptive decoding component, a parallel two-layer feature pyramid structure extracts and fuses low-level and high-level feature information, ensuring the subtle features of object edges are fully utilized in predicting the area, angle, and width of each potential grasping point of the drug.

However, the AFFGA-Net can suffer from degraded grasp detection performance due to high computational costs and background interference in complex visual environments. To address this, the input image is pre-processed with YOLO-EASB, effectively filtering out non-essential background noise and retaining only the key contour information of the drug to be grasped. This step significantly reduces the redundant data processed by the grasping network, saving computational resources and improving overall efficiency.

In the drug grasping scenario, the proportion of effective grasping points in the visual input is extremely limited. The traditional Binary Cross-Entropy (BCE) loss function used in the AFFGA-Net may not focus sufficiently on difficult points during training. Therefore, BCE is replaced with the Focal Loss (FL) function\citep{lin2017focal}. Specifically, Focal Loss is defined as follows:
\begin{equation}
\mathrm{FL}\left(\mathrm{p}_{\mathrm{t}}\right)=-\alpha_{\mathrm{t}}\left(1-\mathrm{p}_{\mathrm{t}}\right)^\gamma \log \left(\mathrm{p}_{\mathrm{t}}\right)
\end{equation}
In the equation for FL, $\mathrm{p}_{\mathrm{t}}$ represents the predicted probability of the correct category for the sample. The term $-\alpha_{\mathrm{t}}$ is the balancing factor, which adjusts the imbalance between positive and negative samples. The term $\left(1-\mathrm{p}_{\mathrm{t}}\right)$ is the error term, reducing the weight of easily classified samples and increasing the weight of difficult ones. The parameter $\gamma$ is the focusing parameter, adjusting the loss weight of easily classified samples. Finally, $\log \left(\mathrm{p}_{\mathrm{t}}\right)$ is the logarithmic loss term, measuring the gap between the predicted probability and the actual label.

\subsection{Framework for robotic arm trajectory planning}
\begin{figure}[h]
    \centering               
\includegraphics[width=.5\textwidth]{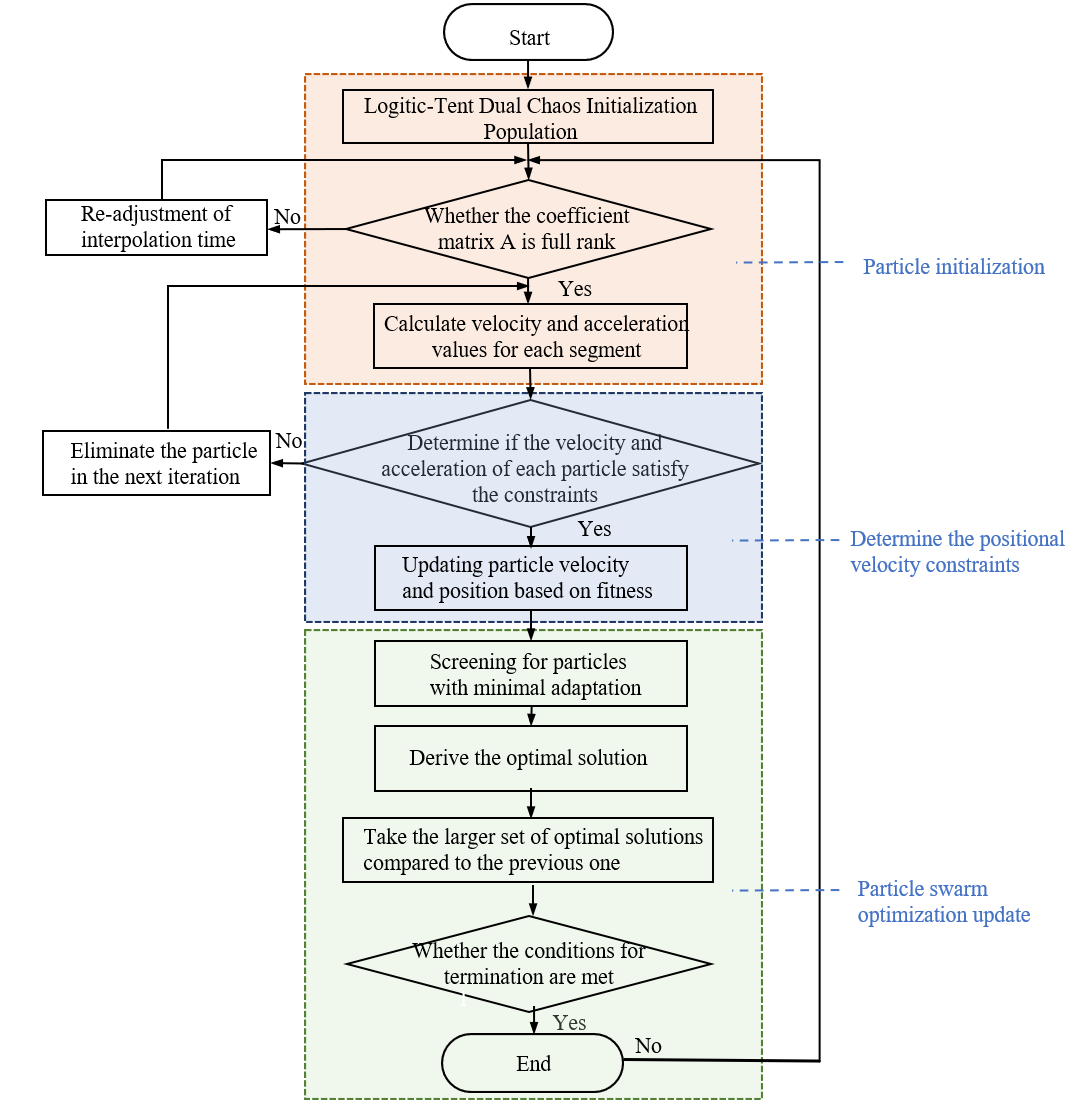}
    \caption{Flow of time-optimal algorithm for trajectory planning}
    \label{fig.8}
\end{figure}

In order to improve the smoothness and stability of the robotic arm's gripping trajectory while optimizing the operation time, this paper proposes a trajectory planning method that optimizes the interpolation time of 3-5-3 segmented polynomials using an improved PSO particle swarm algorithm. This method aims to meet the operating speed and acceleration constraints of the robotic arm \citep{jiang2022time}. Firstly, a 3-5-3 segmented polynomial interpolation function is constructed. Then, an objective function is formulated based on the interpolation times of the segments. Finally, under predefined constraints, the interpolation times are optimized using the improved PSO particle swarm algorithm to minimize the operation time of the robotic arm. The algorithm flow is illustrated in Figure \ref{fig.8}.

\textbf{a. Improved particle swarm algorithm}\\
When using the standard particle swarm algorithm \citep{marini2015particle} to optimize the problem, its parameters are fixed. The inertia weight of the particles, which controls their movement, should be larger in the early stages of optimization to ensure that each particle can explore the search space independently. Later, it should be smaller to allow particles to converge towards better solutions found by other particles. The maximum step size for individual and global particles, \( \psi \) and \( \phi \) respectively, should be larger initially to balance global and local search abilities, and smaller later to refine local solutions. These parameters affect the flight direction of particles. If the inertia is too high, particles may not adjust to better positions found by others, potentially causing premature convergence or slow convergence in later stages of the algorithm.

The improved particle swarm algorithm proposes a nonlinear decreasing strategy to adjust the value of the inertia weight \( \omega \) to address the issue of premature convergence or slow convergence in the later stages of the standard particle swarm algorithm. Specifically, the expression is:
  
\begin{equation}
\omega=\omega_{\min }+\frac{\omega_{\max }-\omega_{\min }}{2}\left[1+\cos \frac{(n-1) \pi}{N-1}\right]
\end{equation}
where $\omega_{\max }$, $\omega_{\min }$ are the maximum and minimum inertia weights, respectively, and in this paper, 0.86 and 0.44 are taken. $n$ is the current iteration number; $N$ is the maximum number of evolutionary generations.
And the method of adopting dynamics is used to assign values to $c_1$ and $c_2$:
\begin{equation}
c_1=c_{11}+\sin \left[\frac{\pi}{2}\left(1-\frac{n}{N}\right)\right]
\end{equation}

\begin{equation}
c_2=c_{21}-\sin \left[\frac{\pi}{2}\left(1-\frac{n}{N}\right)\right]
\end{equation}

where $c_{11}$, $c_{21}$ are constants, and in this paper  $c_{11}$ and $c_{21}$ are taken as 1.3.

However, IPSO initiates the primary stage by randomly generating \( M \) three-dimensional particles, which can lead to uneven distribution of the population and affect the algorithm's optimization. Therefore, this paper introduces an initialization sequence based on the Logit-Tent double-mixing degree approach, building on IPSO \citep{jiang2007improved}. Logistic mapping (taking \( \mu = t \)) is employed to generate \( i \) chaotic sequences \( chx_{\eta} \), where \( t = 1, 2, \cdots, D \) and \( \eta = 1, 2, \cdots, m \). The chaotic sequence \( chx \) is then transformed into optimized variables through a carrier transformation process.
\begin{equation}
x_{t j}=x_{\max , \mathrm{J}}-\left(x_{\max , \mathrm{J}}-x_{\operatorname{mn}, \mathrm{J}}\right) \cdot \operatorname{ch} x_{t J}
\end{equation}

Where \( x_{\text{max}} \) and \( x_{\text{min}} \) denote the maximum and minimum values of the optimization variable \( x \), \( D \) is the variable dimension, and \( m \) is the population size. The mapped optimization variable \( x_1 \) is used as the initialization value of the particle. The fitness of the particles is calculated, and then a rough search for particles is conducted. For particles that mature early, chaotic perturbation is applied using the Tent mapping (with \( \varphi = 0.6 \)) to generate the sequence \( thx_v \), where \( i = 1, 2, \ldots, D \) and \( j = 1, 2, \ldots, m \). The chaotic perturbation is performed using the following equation:

\begin{equation}
\psi^{+}=\left(P_{d}-\alpha \cdot \operatorname{th} x_{t j}\right) /(1-\alpha)
\end{equation}
\begin{equation}
\mathrm{I}^{* *}=\psi^*\left(x_{\max , \mathrm{j}}-x_{\min , \mathrm{J}}\right)+x_{\text {mun } \mathrm{J}}
\end{equation}

where $\alpha(0 \leq \alpha \leq 1)$ is the tuning parameter, $P_{d}(d=1,2 \cdots D)$ is the current optimal solution vector; where $\mathrm{I}^*=\left(x_1^*, x_2^* \cdots x_b^*\right)$ is the chaotic vector after perturbation. 

Thus, the particles can search the whole space on the basis of fast local optimization, which effectively improves the accuracy and convergence speed of the particle swarm algorithm.

In the Method section, a multi-stage grasping framework is introduced. An improved SRCNN model is used for super-resolution reconstruction to enhance image quality and feature accuracy. The YOLO-EASB model is employed for precise drug instance segmentation, integrating ECA and ADaFF modules to improve recognition across scales. The robot arm's trajectory is optimized using an improved particle swarm algorithm, ensuring smoother and more efficient grasping. This framework demonstrates high accuracy and adaptability in complex drug stacking and occlusion scenarios.

\section{Experiment}
This section presents the experimental setup and results for validating the proposed multi-stage grasping framework. The experiments include drug selection, dataset preparation, and performance evaluation. Key metrics such as accuracy, precision, and recall are measured, along with the performance of the robotic arm's trajectory planning and grasp detection networks. The results demonstrate the effectiveness of the system in handling complex drug sorting tasks, showcasing significant improvements in grasping accuracy, efficiency, and adaptability under real-world conditions.
\subsection{Experiment-setup}
\begin{figure}[h]
    \centering               
\includegraphics[width=0.5\textwidth]{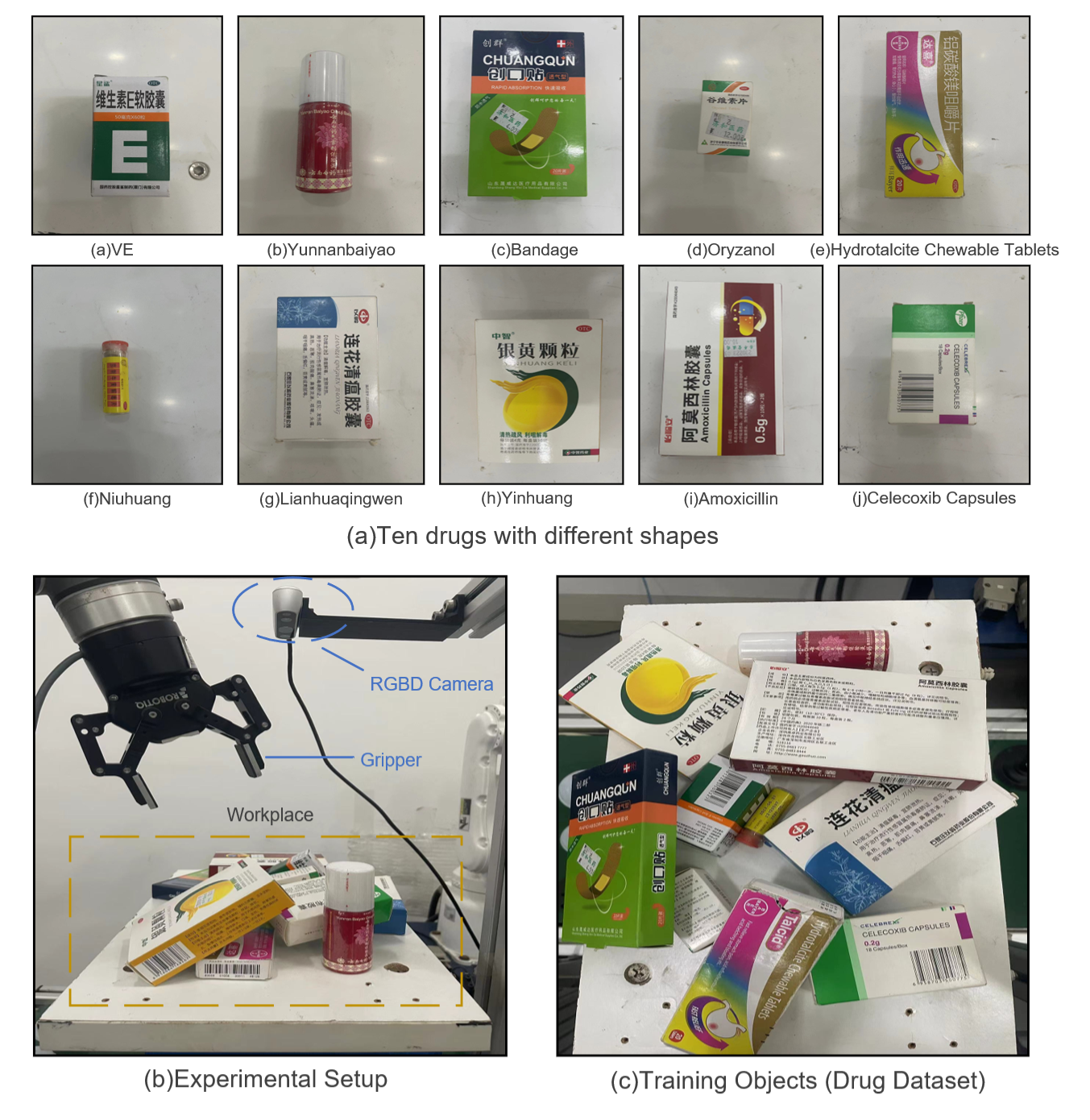}
    \caption{Physical platform building(a)Ten drugs with different shapes (b)Experimental Setup (c) Training Objects(Drug Dataset)}
    \label{fig.9}
\end{figure}

\textbf{a. Drug selection and physical platform construction}\\
In the actual experiment, 10 common medicines from a pharmacy located in China were selected to evaluate the model's performance, as shown in Figure \ref{fig.9}(a). These medicines include Yunnan Baiyao, Band-Aid, Oryzanol, Hydrotalcite Chewable Tablets, Niuhuang detoxification granules, Lotus capsules, Yinhuang granules, Amoxicillin, vitamin E capsules and Celecoxib Capsules. The drug shapes vary: several rectangular shapes with different lengths, widths, and heights, and cylinders with different diameters and heights. The filming location is in Quanzhou, Fujian, China (latitude 24$^\circ$52$^\prime$32.0$^\prime\prime$ N, longitude 118$^\circ$ 40$^\prime$ 20.5$^\prime\prime$  E). Drug images were captured using a D435 camera installed on the experimental drug sorting platform. The initial size of the images is 640 pixels × 480 pixels. A total of 600 images were collected for the dataset, and the image format is RGB.

Rectangular drugs, due to their regular shape, are easy to stack, with their edges and surfaces in close contact. This can cause occlusion, making edge detection difficult, especially when the colors and textures are similar. Occluded parts may be misidentified as background or other medicines. Irregular occlusions in both rectangular and cylindrical drugs lead to complex edge detection and shape recognition problems. Cylindrical drugs, due to their curved surfaces, may have occlusions recognized as part of a curve. When stacked, they may roll, making their stacking less stable than rectangular drugs. The physical experimental platform is shown in Figure \ref{fig.9}(b). The medicine images are acquired by a Realsense D435 camera fixed above the medicines, and a two-finger gripping claw is fitted at the end of a UR5 robotic arm for grasping. The medicines are placed as shown in Figure \ref{fig.9}(c), overlapping each other in different postures.

\textbf{b. Self-constructed drug dataset}

\begin{figure}
	\centering
	\begin{minipage}{0.45\linewidth}
		\centering
		\includegraphics[width=1\linewidth]{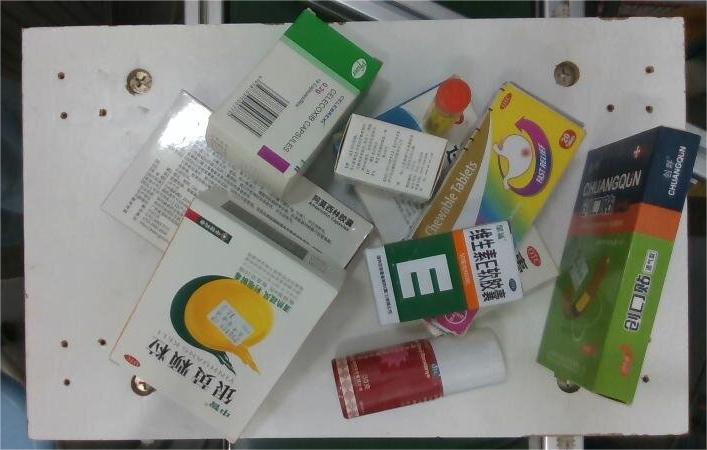}
            \subcaption{(a)dataset}
		
	\end{minipage}
	\begin{minipage}{0.45\linewidth}
		\centering
		\includegraphics[width=1\linewidth]{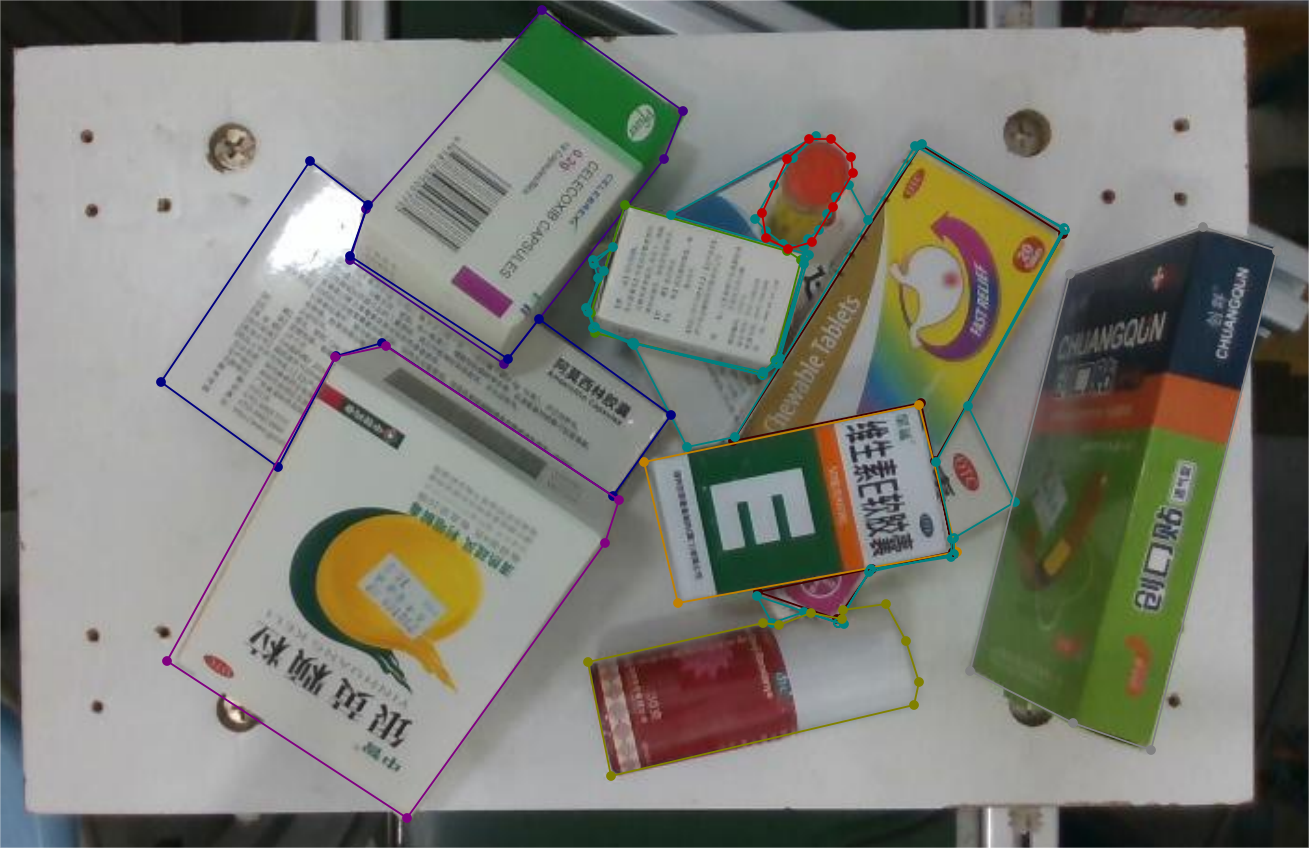}
            \subcaption{(b)labelme label}
  
	\end{minipage}
 \caption{Visualization of an example in self-constructed drug dataset}
    \label{fig.10}
\end{figure}

The drug testing dataset used in this task is a custom dataset. It consists of a series of images with dimensions of 640 pixels × 480 pixels captured using a D435 camera installed on the experimental drug sorting platform. As depicted in Figure \ref{fig.10}, each image contains between 1 to 10 stacked objects. The dataset comprises a total of 600 images labeled using LabelMe. The dataset is split in a 7:2:1 ratio, 420 with images allocated for the training set and 120 images for the test set,60 images for the validation set.\\
\textbf{c. Experimental Configuration}

This study utilized the Ubuntu 20.04 operating system, an Intel(R) Core(TM) i5-8500 CPU, 16GB of RAM, and an NVIDIA GeForce RTX 3060 GPU with 8GB of video memory. Python 3.8 and PyTorch 1.9.1 were used as the Deep Learning framework. Drug names in the experimental dataset were abbreviated by replacing the first three letters of each drug: Yun (Yun Nan Bai Yao), Ban (Band-Aid,), Ory (Oryzanol), Hyd (Hydrotalcite Chewable Tablets), Niu (Niu Huang Detoxification Granules), Lia (Lotus Capsules), Yin (Yin Huang Granules), Amo (Amoxicillin), VE(Vitamin E Capsules) and Cel (Celecoxib Capsules).

\subsection{Instance segmentation experiment}

\textbf{a. Comparison experiment}

To further validate the model performance, YOLO-EASB model based on improved YOLOv5 is tested against mainstream instance segmentation algorithm models such as YOLOACT \citep{bolya2019yolact}, SOLOv2 \citep{wang2020solov2}\\
YOLOv7-seg \citep{wang2023yolov7}, and YOLOv8-seg \citep{dumitriu2023rip} and Mask-RCNN \citep{he2017mask}  in the same environment, and the results are shown in Table 1 and Table 2. From the table, it is evident that the proposed method in this chapter outperforms other methods in terms of mAP50, Precision, and Recall metrics. It achieves higher Precision values of 2.7\% and 4\%, and higher Recall values of 2.9\% and 2.5\% than the existing popular models.
\begin{table*}[t!]
\caption{Yolo-Seg improves the comparative experiment}\label{biao4_34}
\begin{tabular}{lcccccccccccc}
\hline
\multicolumn{1}{c}{}                                  & \multicolumn{10}{c}{\textbf{Mean average precision mAP50/\%}}                                                                                                                                                                                                                                                                                                                                       &                                         & \multicolumn{1}{l}{}                                     \\ \cline{2-11}
\multicolumn{1}{c}{\multirow{-2}{*}{\textbf{Models}}} & \multicolumn{1}{l}{Yun}              & \multicolumn{1}{l}{Amo}              & \multicolumn{1}{l}{Yin}              & \multicolumn{1}{l}{Lia}              & \multicolumn{1}{l}{Ban}              & \multicolumn{1}{l}{Hyd}              & \multicolumn{1}{l}{Niu}              & \multicolumn{1}{l}{Ory}              & \multicolumn{1}{l}{vE}               & \multicolumn{1}{l}{Cel}              & \multirow{-2}{*}{\textbf{Precision/\%}} & \multicolumn{1}{l}{\multirow{-2}{*}{\textbf{Recall/\%}}} \\ \hline
YOLACT                                                & 95.3                                 & 96.8                                 & 97.2                                 & 92.3                                 & 92.1                                 & 85.3                                 & 88.3                                 & 91.7                                 & 95.1                                 & 92.1                                 & 93.2                                    & 88.3                                                     \\
SOLOv2                                                & 96.1                                 & 98                                   & 96.9                                 & 94.6                                 & 95.3                                 & 89.2                                 & 92.3                                 & 91.8                                 & 93.5                                 & 94.2                                 & 91.4                                    & 92.3                                                     \\
YOLOv7X-seg                                           & 96.8                                 & 97.9                                 & 98.1                                 & 95.2                                 & 95.6                                 & 85.8                                 & 89.5                                 & 92.4                                 & 91.4                                 & 93.7                                 & 94.8                                    & 92.7                                                     \\
{YOLOv8X-seg}                    & {97.1}          & {97.6}          & { 97.9}          & {95.5}          & {96.8}          & {88.9}          & {91.8}          & {93.3}          & {92.8}          & {95.4}          & { 93.5}             & {95.2}                              \\
Mask-RCNN                                             & 92.4                                 & 95.7                                 & 96.1                                 & 91.4                                 & 91.8                                 & \multicolumn{1}{l}{85.5}             & 87.4                                 & 89.2                                 & \multicolumn{1}{l}{90.5}             & \multicolumn{1}{l}{92.4}             & 90.8                                    & 87.8                                                     \\
{\textbf{YOLOv5-EASB}}           & {\textbf{99.3}} & {\textbf{99.5}} & {\textbf{99.4}} & {\textbf{98.3}} & {\textbf{98.1}} & {\textbf{93.8}} & { \textbf{95.4}} & {\textbf{96.6}} & {\textbf{95.8}} & {\textbf{99.2}} & { \textbf{97.5}}    & {\textbf{95.2}}                     \\ \hline
\end{tabular}
\end{table*}

YOLOv7X-seg and YOLOv8X-seg, respectively. Compared with the classical models Mask-RCNN, the Precision value is higher by 6.7\% and the Recall value is higher by 7.4\%. Compared with the single-stage methods YOLACT and SOLOv2, the Precision of the proposed model in this chapter is higher by 4.3\% and 6.1\%, respectively. These results demonstrate that the YOLO-EASB model in this paper achieves good segmentation performance. However, the mAP is only 95.8\% in the segmentation of drugs with small targets such as VE, indicating a direction for future improvement of the model.

\textbf{b. Ablation experiment}

In this subsection, ablation studies are conducted to evaluate the impact of each component of the proposed YOLO-EASB on performance. All networks are trained and tested on a self-constructed drug dataset. The results are summarized in Table. Incorporating SPPCFCSPC and BiFPNC improves mAP by different margins. Specifically, YOLOv5+E-A-SPPCFCSPC and YOLOv5+BiFPNC enhance pharmaceutical segmentation accuracy by 3.4\% and 3.3\% , respectively, compared to YOLOv5. The proposed enhancements improve pharmaceutical segmentation accuracy by 3.7\% over YOLOv5, demonstrating that YOLO-EASB enhances instance segmentation capability in scenarios where drugs occlude each other.
\begin{table*}
\caption{Ablation experiments related to YOLO-seg performance improvement}\label{biao4_35}
\begin{tabular}{lclcccccccccc}
\hline
\multicolumn{1}{c}{}                                  & \multicolumn{10}{c}{\textbf{Mean average precision mAP50/\%}}                                                                                                                                                                                                                                                                                                                                                           &                                         & \multicolumn{1}{l}{}                                     \\ \cline{2-11}
\multicolumn{1}{c}{\multirow{-2}{*}{\textbf{Models}}} & \multicolumn{1}{l}{Yun}              & Amo                                  & \multicolumn{1}{l}{Yin}              & \multicolumn{1}{l}{Lia}              & \multicolumn{1}{l}{Ban}              & \multicolumn{1}{l}{Hyd}              & \multicolumn{1}{l}{Niu}                                  & \multicolumn{1}{l}{Ory}              & \multicolumn{1}{l}{VE}               & \multicolumn{1}{l}{Cel}              & \multirow{-2}{*}{\textbf{Precision/\%}} & \multicolumn{1}{l}{\multirow{-2}{*}{\textbf{Recall/\%}}} \\ \hline
YOLOv5                                                & 96.4                                 & \multicolumn{1}{c}{97.3}             & 97.2                                 & 89.1                                 & 92.8                                 & 83.7                                 & 88.3                                                     & 95.7                                 & 91.1                                 & 93.1                                 & 93.8                                    & 87.1                                                     \\
YOLOv5+E-A-SPPCFCSPC                                  & 98.7                                 & 99.2                                 & 98.1                                 & 94.2                                 & 95.8                                 & 93.7                                 & 95.1                                                     & 96.6                                 & 95.3                                 &98.6                                   & 97.2                                    & 94.1                                                     \\
YOLOv5+BIFPNC                                          & 96.8                                 & 99.4                                 & 97.8                                 & 95.4                                 & 97.3                                 & 92.5                                 & 94.6                                                     & 96.5                                 & 94.5                                 & 99.1                                 & 97.1                                    & 94.2                                                     \\
{\textbf{YOLO-EASB}}           & {\textbf{99.3}} & {\textbf{99.5}} & {\textbf{99.4}} & {\textbf{98.3}} & {\textbf{98.1}} & { \textbf{93.8}} & \multicolumn{1}{l}{{\textbf{95.4}}} & {\textbf{96.6}} & {\textbf{95.8}} & {\textbf{99.2}} & { \textbf{97.5}}    & { \textbf{95.2}}    \\ \hline                
\end{tabular}
\end{table*}
 As shown in Tables 1 and 2, YOLO-EASB has higher Recall, mAP, and P than the instance segmentation algorithm that incorporates separate modules.\\
\begin{figure*}
    \centering               
\includegraphics[width=0.99\textwidth]{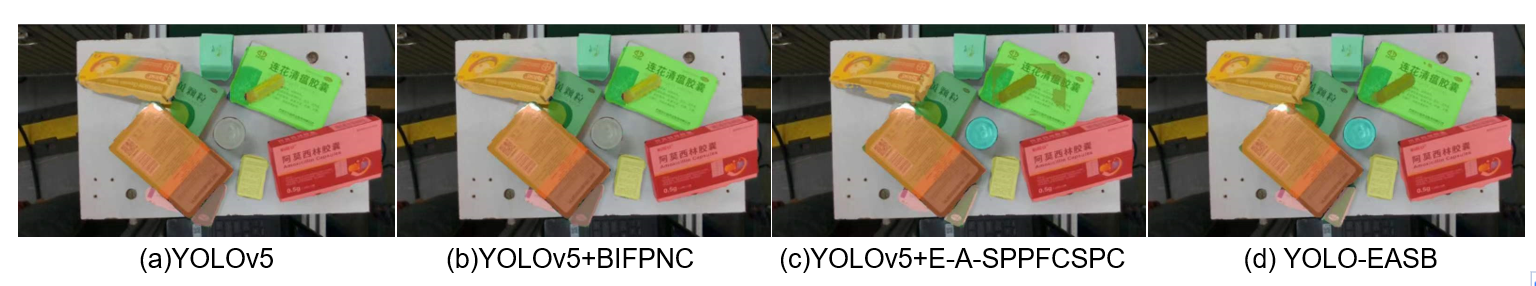}
     \caption{The detection performance based on a sample using (a) YOLOv5; (b) YOLOv5+BIPFNC; (c) YOLOv5+E-A-SPPCFCSPC; (d) YOLO-EASB}
    \label{fig.11}
\end{figure*}

The results are visualized in Figure \ref{fig.11}(a)(b). For the cylindrical drug Yunnan Baiyao, both YOLOv5 and YOLOv5+BIFPNC failed to recognize and segment the target due to its cylindrical surface being close to white and similar to the background color, which interferes with the model's recognition, leading to detection omission. The segmentation results of the yolov5+E-A-SPPFCSPC model are shown in Figure \ref{fig.11}(c); while Yunnan Baiyao is detected, the segmentation of the drug Daxi shows uneven edge segmentation. Figure \ref{fig.11}(d) illustrates that the YOLO-EASB model can effectively perform instance segmentation and clearly segment all the drugs. Through visual verification, the YOLO-EASB model proves significantly more effective for images containing overlapping and visually similar medicines.\\

\subsection{Grasp Detection Network Experiment}
In Table 3, IAFFGA-Net is compared with representative planar grabbing detection methods in the public Cornell dataset under the same experimental conditions. The dataset is a self-constructed drug dataset, and the data input formats of the selected grabbing detection methods are all in the same RGB format as the methods in this paper. Focal Loss parameters are $\alpha$= 0.25 and $\Gamma$ = 2.
\begin{table*}
\caption{Comparative Experiments on Grabbing Networks}\label{biao4_31}
\begin{tabular}{ccccccc}
\hline
\textbf{Method}     & \multicolumn{1}{l}{\citep{asif2017rgb}} & \multicolumn{1}{l}{\citep{song2020novel}} & \multicolumn{1}{l}{\citep{asif2019densely}} & \multicolumn{1}{l}{\citep{zhou2018fully}} & \multicolumn{1}{l}{AFFGA} & \multicolumn{1}{l}{ \textbf{Ours}} \\ \hline
\textbf{Prediction(\%)}    & 84.1                            & 85.6                            & 86.3                            & 87.4                            & 88.7                      & \textbf{90.2 }                      \\
\textbf{speed(fps)} & -                               & -                               & 9.1                             & 8.6                             & 55.5                      & \textbf{57.2 }                      \\ \hline
\end{tabular}
\end{table*}
The results show that our proposed IAFFGA-Net is more accurate  and the detection speed is slightly faster in chaotic overlapping scenes compared to the other grasping networks.

\subsection{Performance of Multi-stage Grasp Framework}
To further validate the Multi-stage  grasping framework proposed in this paper, performance tests were conducted for IAFFGA, YOLO-EASB+IAFFGA, and SRCNN+YOLO-EASB+IAFFGA, respectively. The test results are shown in Table 4. As shown in Table 4, the accuracy of the Multi-stage grasping framework reaches 97.3\% on the dataset images. Although there is an increase in running time, it still meets the real-time demand for the smart pharmacy drug dispensing scenario discussed in this paper.
\begin{table}[h]
\caption{Multi-stage Grasp Framework Performance Experiment}\label{biao4_32}
\centering 
\begin{tabular}{ccc}
\hline
\textbf{Method}  & \multicolumn{1}{l}{\textbf{Predict(\%)}}  & \multicolumn{1}{l}{\textbf{time/s}} \\ \hline
IAFFGA                               & 90.2                                                                             & 0.23                                \\
YOLO-EASB+IAFFGA                              & 96.6                                                                             & 0.41                                \\ \hline
\textbf{ SRCNN+YOLO-EASB+IAFFGA } &  \textbf{97.3 }                              &\textbf{ 0.45 }       \\ \hline
\end{tabular}
\end{table}
\subsection{Robotic arm trajectory planning simulation experiment}
\begin{figure*}[t!]
    \centering               
\includegraphics[width=1\textwidth]{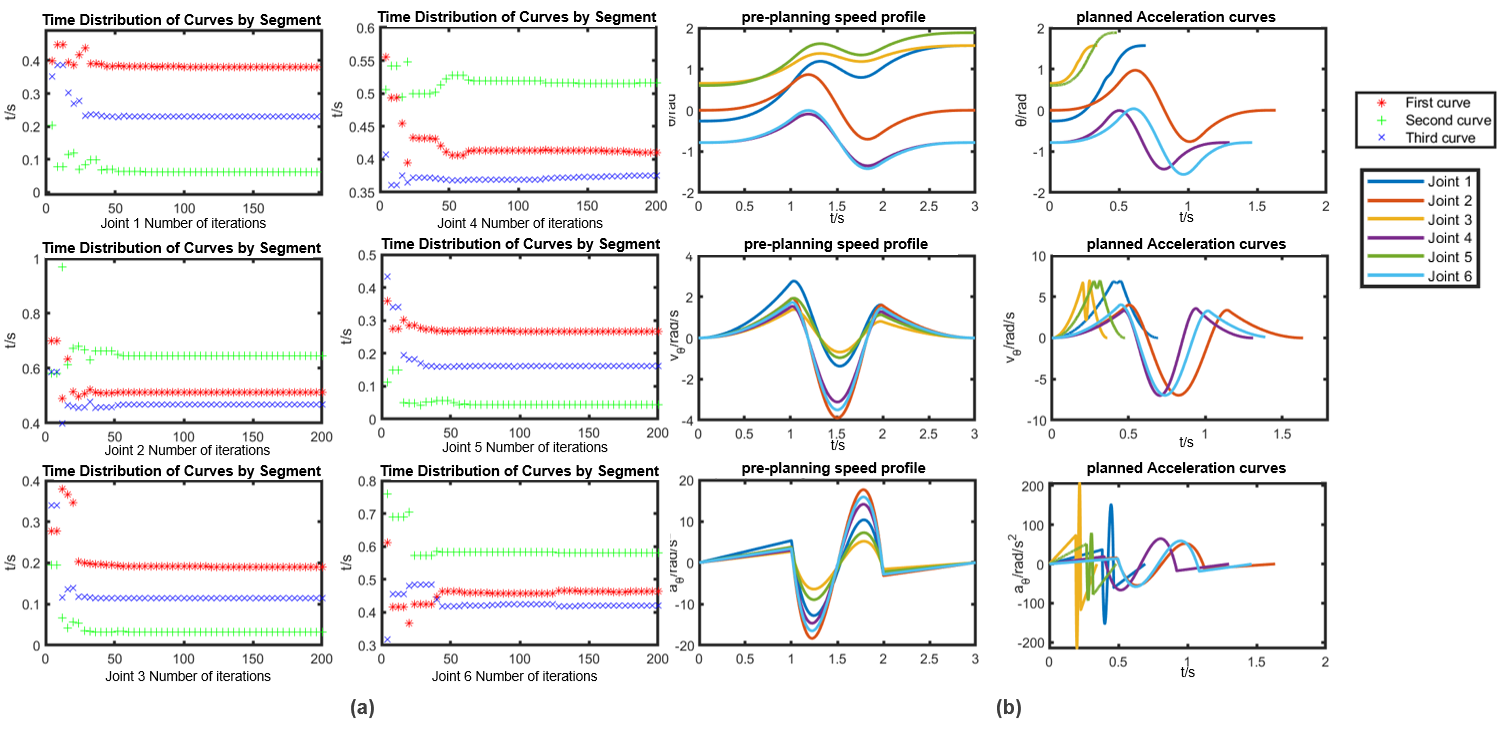}
    \caption{(a) Particle position evolution curve of joints 1-8; (b) Comparison dlagram of the Posltlon, Speed Proflle, and Acceleratlon of the planned front and rear robotic arms.}
    \label{fig.12}
\end{figure*}

During the experiment, the improved particle swarm algorithm achieved asymptotic convergence in approximately 25 iterations, while the standard particle swarm algorithm requires approximately 55 iterations. This indicates that the efficiency of the improved particle swarm algorithm has increased by 45\%. Figure \ref {fig.12} (a) shows the optimized 3-stage interpolation time values for each joint of the robot arm obtained by solving the time optimization problem.

Meanwhile, a comparison of the robotic arm's position, velocity, and acceleration before and after planning is presented, as shown in Figure \ref{fig.12}(b). It can be observed that the displacement, velocity, and acceleration curves of the robotic arm under the optimization of the improved particle swarm algorithm are continuous and free of sudden changes, indicating smooth operation through each path point. Moreover, the velocity and acceleration of each joint of the robotic arm meet the specified constraints. Concurrently, the time is reduced by 1.1 seconds, significantly enhancing the grasping efficiency of the system.

\subsection{Actual system grabbing experiments}
\begin{figure}[h]
    \centering               
\includegraphics[width=0.5\textwidth]{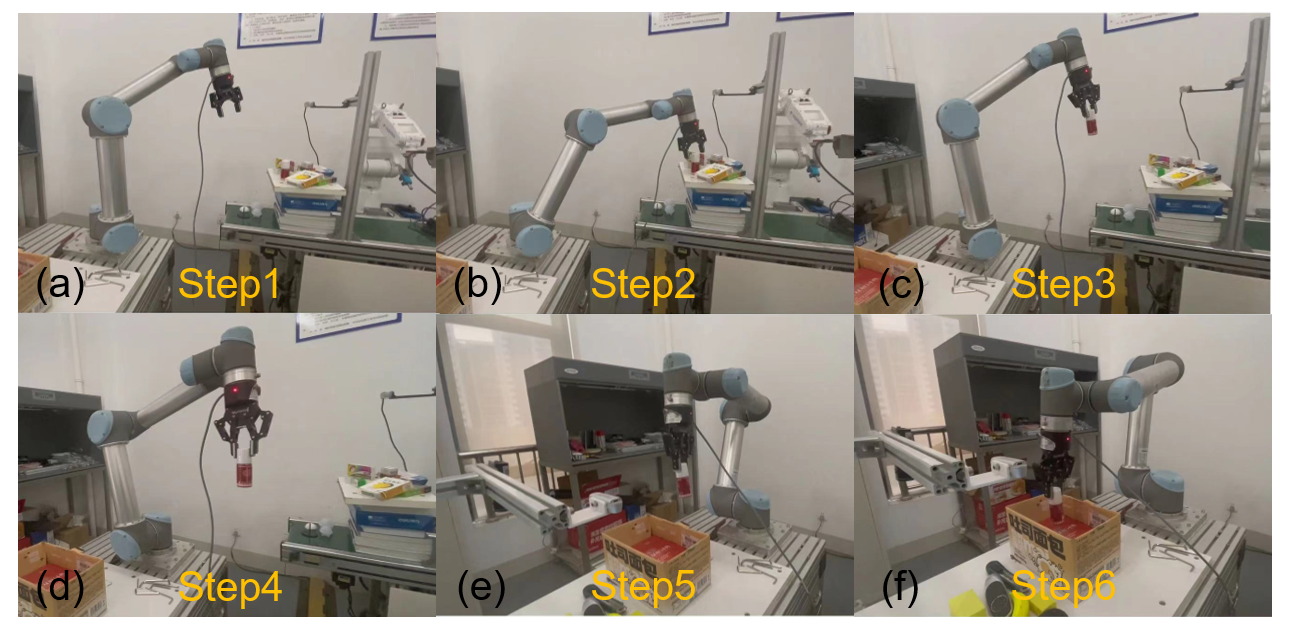}
    \caption{Physical experiment (a) Step1: start from the initial position (b) Step2: grab the medicine (c) Step3: after grabbing the medicine, adjust the position to the appropriate location (d) Step4: start trajectory planning (e) Step5: trajectory planning to move to the medicine distribution place (f) Step6: move to the distribution place to be placed in order. The full capture video can be viewed at https://tomry1114.github.io/robot-grasp/}
    \label{fig.13}
\end{figure}

This paper develops a robotic arm gripping and inspection system comprising a UR5 robotic arm, a two-finger mechanical gripper, and a D435 depth camera. A D435 depth camera is mounted on the upper part of the table to capture high-quality RGB-D information. The experimental setup is illustrated in Figure \ref{fig.13}, where the robotic arm initiates from an initial position, receives positional information from the camera sensor, moves to the drug sorting area for grasping, and subsequently transfers to the drug dispensing box for orderly arrangement.\\
Ten different position placements were gone through, and the number of times each drug was successfully grabbed was recorded. And compared with the AFFGA-Net, the actual grasping experiment results are shown in Table 5.\\

\begin{table*}
\caption{The performance based on actual gripping experiments}\label{biao4_33}
\centering
\begin{tabular}{cllllllllll}
\midrule[0.5mm]
\textbf{drug}               & \textbf{vE}                    & \textbf{Yun}                   & \textbf{Ban}                   & \textbf{Ory}                   & \textbf{Hyd}                   & \textbf{Niu}                   & \textbf{Lia}                   & \textbf{Yin} & \textbf{Amo} & \textbf{Cel} \\ \cline{2-11} 
\textbf{success proportion} & \multicolumn{1}{c}{10/10}      & \multicolumn{1}{c}{9/10}       & \multicolumn{1}{c}{10/10}      & \multicolumn{1}{c}{10/10}      & \multicolumn{1}{c}{10/10}      & \multicolumn{1}{c}{8/10}       & \multicolumn{1}{c}{9/10}       & 10/10        & 10/10         & 9/10         \\ \midrule[0.5mm]
\textbf{experiment num}     & \multicolumn{1}{c}{\textbf{1}} & \multicolumn{1}{c}{\textbf{2}} & \multicolumn{1}{c}{\textbf{3}} & \multicolumn{1}{c}{\textbf{4}} & \multicolumn{1}{c}{\textbf{5}} & \multicolumn{1}{c}{\textbf{6}} & \multicolumn{1}{c}{\textbf{7}} & \textbf{8}   & \textbf{9}   & \textbf{10}  \\ \cline{2-11} 
\textbf{time/min}           & 1.56                           & 1.15                           & 1.12                           & 1.25                           & 2.10                           & 2.22                           & 2.38                           & 1.45         & 2.06         & 1.11         \\ \midrule[0.5mm]
\end{tabular}
\end{table*}
The table indicates that in real grasping experiments, vE, Ban, Ory, Hyd, Yin, and Amo achieved a 100\% success rate, while Yun, Lia, and Cel experienced one grasping failure out of ten attempts. Additionally, there were two grasping failures for smaller cylindrical drugs, Niu, highlighting the need to improve the performance of our grasping detection network for small targets and cylindrical shapes, which remains a goal for future work. In laboratory testing, the robotic arm is constrained to a speed limit of 250 mm/sec for safety. The highest time taken in ten experiments to grasp ten pieces of drugs was 2.38 minutes, meeting real-time operational requirements.
\section{CONCLUSION}
In this paper, an innovative intelligent drug sorting system that integrates advanced software algorithms with hardware functionalities was developed and evaluated to tackle complex grasping challenges in smart pharmacies. A multi-stage grasping framework that utilizes an improved SRCNN model for super-resolution reconstruction of drug images is proposed, improving the accuracy of feature capture and providing clear visual input for grasping decisions. Propose the YOLO-EASB instance segmentation model for high-precision spatial localization and feature extraction of drugs, followed by target drug segmentation, background subtraction, and input into the IAFFGA-net to obtain the grasping angle and width of drugs. In order to solve the problems of smooth robot arm trajectory and execution efficiency, an improved PSO algorithm was used for time optimization, which improved the system's operational capability and ensured smooth and efficient trajectory planning. Experimental results demonstrated the superiority of our multi-stage grasping framework in optimizing smart pharmacy operations. Our system showed significant improvements in accuracy and efficiency compared to existing methods, achieving real-time drug sorting with high precision in complex environments. The proposed system also exhibited remarkable adaptability and effectiveness in practical applications.

Future work will focus on further optimizing the grasping performance for small-target and cylindrical shapes drugs, extending the system to a wider variety of drug types and more complex pharmacy environments. To achieve this, domain adaptation techniques \citep{Qiu2024} will be explored to enhance the model's generalizability across different pharmacy settings, ensuring robust performance even when introduced to new drug types or operational conditions not covered during initial training. Additionally, improvements in real-time processing capabilities through hardware acceleration and algorithmic optimizations will be pursued, alongside exploring human-robot collaboration to enhance system flexibility and intelligence.


\section{Acknowledgements}
This work was supported by grants from the National Key Research and Development Program of China (Grant No. 2022YFF0710800), Major International (Regional) Joint Research Project of China (Grant No. 81820108001).

\bibliographystyle{elsarticle-harv} 
\bibliography{example}






\end{document}